\documentclass[a4paper, 10pt, conference]{IEEEtran}

\IEEEoverridecommandlockouts

\usepackage{cite}
\usepackage{amsmath,amssymb,amsfonts}
\usepackage{algorithmic}
\usepackage{graphicx}
\usepackage{textcomp}
\usepackage{xcolor}
\usepackage{booktabs}
\usepackage{multirow}
\usepackage{amssymb}
\usepackage{float}
\usepackage{url}
\usepackage{balance}
\usepackage{xspace}
\usepackage[ruled]{algorithm2e}
\usepackage{balance}

\def\BibTeX{{\rm B\kern-.05em{\sc i\kern-.025em b}\kern-.08em
    T\kern-.1667em\lower.7ex\hbox{E}\kern-.125emX}}

\usepackage{soul}

\begin{document}

\title{
Pattern Spotting and Image Retrieval in Historical Documents using Deep Hashing \\
}

\newif\ifauthors
\authorstrue

\ifauthors
\author{
    \IEEEauthorblockN{
        Caio da S. Dias\textsuperscript{1}, 
        Alceu de S. Britto Jr\textsuperscript{1,4}, 
        Jean P. Barddal\textsuperscript{1},
        Laurent Heutte\textsuperscript{3},
        Alessandro L. Koerich\textsuperscript{2}
    }
    \IEEEauthorblockA{
        \\ 
        \textsuperscript{1}Graduate Program in Informatics (PPGIa), Pontifícia Universidade Católica do Paraná (PUCPR), Curitiba (PR), Brazil
        \\
        Email: \{caio.dias, alceu, jean.barddal\}@ppgia.pucpr.br
        \\
        \textsuperscript{3}Normandie Univ, UNIROUEN, UNIHAVRE, INSA Rouen, LITIS, Rouen, France
        \\
        Email: laurent.heutte@univ-rouen.fr
        \\
        \textsuperscript{2}École de Technologie Supérieure (ÉTS), Université du Québec, Montréal (QC), Canada
        \\
        Email: alessandro.koerich@etsmtl.ca
        \\
        \textsuperscript{4}State University of Ponta Grossa (UEPG), Ponta Grossa (PR), Brazil 
    }
}
\else 
\author{
    \IEEEauthorblockN{Anonymous Authors}
    \vspace*{3.5cm}
}
\fi

\maketitle

\begin{abstract}
This paper presents a deep learning approach for image retrieval and pattern spotting in digital collections of historical documents. 
First, a region proposal algorithm detects object candidates in the document page images. 
Next, deep learning models are used for feature extraction, considering two distinct variants, which provide either real-valued or binary code representations. 
Finally, candidate images are ranked by computing the feature similarity with a given input query. 
A robust experimental protocol evaluates the proposed approach considering each representation scheme (real-valued and binary code) on the DocExplore image database. 
The experimental results show that the proposed deep models compare favorably to the state-of-the-art image retrieval approaches for images of historical documents, outperforming other deep models by 2.56 percentage points using the same techniques for pattern spotting.
Besides, the proposed approach also reduces the search time up to 200$\times$, and the storage cost up to 6,000$\times$ when compared to related works based on real-valued representations.
\end{abstract}


\section{Introduction}

Content-based image retrieval (CBIR), in particular the tasks of image retrieval (IR) and pattern spotting (PS), quickly evolved in recent years and has become essential in computer vision. 
IR aims at retrieving a set of images containing a given search image (query) in their content. 
A search is performed in an image database for each new query, returning the possible candidate images. 
The same occurs in PS, yet, in this task, it is not enough to provide candidate images but also the exact location of the query occurrences (there may be more than one result per image). 
In the context of historical documents, candidate images usually represent images of the document pages.

The massive increase in image collections stored by art museums, medical institutes, and environmental and governmental agencies creates a problem related to information access. 
Often, image indexing is done manually by an individual who filters and inserts a set of keywords in each image to find them easily later.
This process is costly and time-consuming, making indexing large image databases cumbersome. 
An example of this indexing is the digitization of collections of historical documents. 
This task helps ensure that more people have access to its content and assists in securing and preserving the original documents. 
However, as most of these documents were written between the 10th and 16th centuries, continued manipulation can be harmful and even damage these manuscripts. 
Consequently, historians use digitized documents to establish correlations between documents or parts, whether in textual or graphic elements.

Current approaches for indexing rely on automatic search software that increase the efficiency of analyzing large volumes of documents. 
However, with advances in computer vision and machine learning, it is possible to develop applications that find correlations within seconds.

\begin{figure}[htbp]
    \centerline{\includegraphics[scale=0.35]{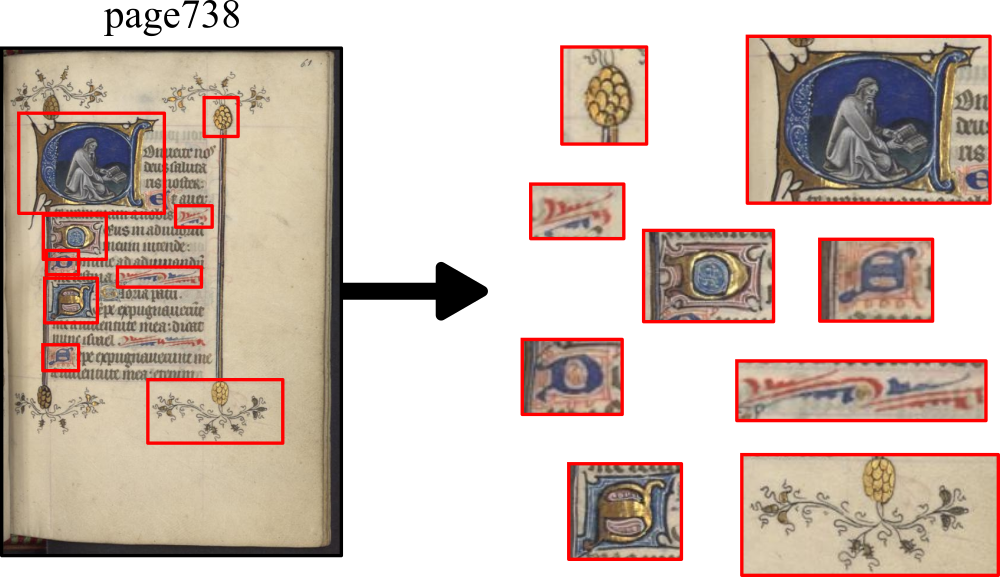}}
    \caption{Example of how a single page of a historical document can contain several figures and special characters to be analyzed.}
    \label{docProblem}
\end{figure}

Historical documents mainly contain handwritten texts but can also have graphical objects \cite{yarlagadda:ACCV:2010}, as shown in Fig.~\ref{docProblem}. 
These graphic objects are special characters, text separators, border details, stamps, coats of arms, and even paintings of festive scenes. 
The main challenge for historians is to establish correlations between different objects in different document collections. 
Among the study topics that these correlations can provide are the characterization of cultural and temporal heritage through patterns of figures and paintings, the categorization of documents by content, and the variability of writing patterns \cite{yarlagadda:ACCV:2010}.



There are several methods of recognizing objects in images, but almost all present two phases: offline and online. An object detector processes the document image files in the offline phase. 
The candidate regions divided into several other files are returned, composing the database of candidate images for the search. 
The processed images are indexed and stored in a predefined structure, informing the image's page, position, and path. 
In the online phase, the similarity measure between the search image and the images of the stored candidate regions is computed. 
These measurements are stored and sorted in ascending order, thus ranking candidate images for each requested search. 
From this similarity ranking, the $n$ smallest distances are selected to compose the result of both IR and PS tasks.

\begin{figure}[ht]
    \centerline{\includegraphics[scale=0.50]{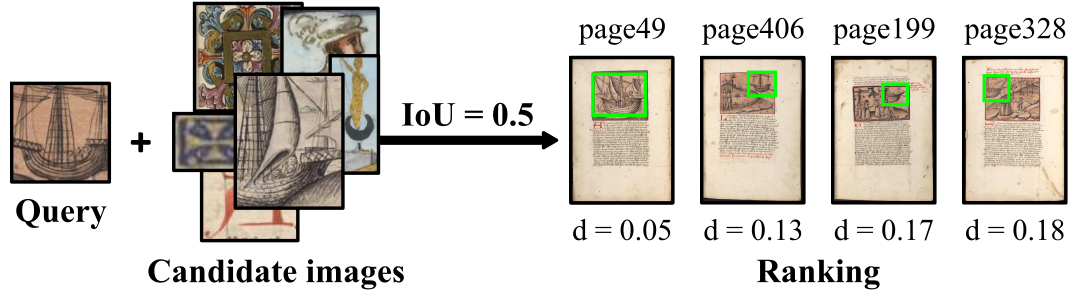}}
    \caption{Searching for an image. The pattern spotting returns a list of non-repeating positions ordered by the distance measure (d), as well as the document page. These positions are subject to minimal overlap between the query image and the processed images measured through the IoU.}
    \label{ps}
\end{figure}

In the IR task, the shortest ranking distances of the Top $n$ candidates are used to return the list of pages where the query can be. 
However, in PS, in addition to returning the pages, the image location within the document is also required. For such an aim, the structure previously stored in the offline phase is used, and it returns this location, as shown in Fig.~\ref{ps} (green rectangles). 
The relevance of a candidate region is related to its overlap with the query, which is given by the intersection over union (\textrm{IoU}) of the query area and the candidate region area. 
For instance, for the analysis of possible candidates, an \textrm{IoU}$\geq$ 0.5 was considered in \cite{docexplore}. 
Precision and recall are calculated, and finally, the mean average precision (mAP) is computed to evaluate the result considering all queries.

This work describes new methods for image retrieval and pattern spotting in historical documents via deep learning techniques, focusing on improving the state-of-the-art considering (i) accuracy, (ii) processing time, and (iii) storage cost. 
For this purpose, real-valued and binary representations generated by various deep model architectures are proposed and evaluated.

The main contribution of this paper is two-fold. 
First, the filtering strategy applied to reduce the number of candidate images (offline phase). 
Second, the binary representation method proposed to reduce the space needed to store the feature maps (offline phase) and the time required to search for a given query (online phase) while improving accuracy compared to related works.

The remainder of this paper is organized as follows. 
Section 2 presents recent works related to IR and PS in historical documents. Section 3 introduces the proposed methods for IR and PS tasks, which employ convolutional neural networks (CNN) and deep hashing. 
It also presents the pre-processing steps and selective search (SS). 
Section 4 presents the database used to evaluate the proposed methods, followed by the experimental protocol, experimental results, and comparison of approaches. 
Finally, conclusions and perspectives for future work are presented in the last section.

\section{Related Works}

Few approaches have been proposed to tackle the problem of PS and IR in historical documents. En \textit{et al.}~ \cite{sovann} proposed a complete system for searching images and locating small graphic objects in medieval documents. Their system is based on the extraction and indexing of regions of interest in the image, representing these regions by handcrafted descriptors and searching for the similarity between query and image candidates using compression and approximation techniques. While this system has shown good performance on \cite{docexplore} medieval document images, it has several weaknesses that make it unsuitable for other types of document images. For instance, it is sensitive to variations in size, shape, color, and patterns to be detected. This system also lacks scaling support and requires post-processing to accurately locate objects in regions of interest using traditional correlation methods.

Ubeda \textit{et al.}~\cite{ignacio} proposed an approach using CNNs based on feature pyramid networks (FPN) as the system feature extractor. FPN allows the extraction of descriptors from localized regions of documents to be indexed in the query at various scales with just one pass through the network. In addition, pre-processing was also used before normalization and FPN application. More specifically, pre-processing regarded background removal from the images in \cite{docexplore} and image centering on a canvas with a black background. The memory requirements and processing time are also reduced when compared to the system proposed in \cite{sovann}.

Another approach using CNN was proposed by Wiggers \textit{et al.}~\cite{kelly}. The approach is based on transfer learning and fine-tuning of a pre-trained CNN. 
It also uses the selective search (SS) algorithm proposed by Uijlings \textit{et al.}~\cite{ss}, which combines segmentation and exhaustive search methods. 
SS was used in the pre-processing step to generate candidate regions within the image dataset. 
The SS consisted of a selective search that combines the strength of an exhaustive search with image segmentation, where the image structure is used to guide the sampling and recognition process. 
As with an exhaustive search, the goal is to capture all possible object locations within the image. SS also uses diversification strategies to combine or differentiate the regions proposed by image segmentation, i.e., color space (C), size (S), texture (T), and fill (F). 
To collect all possible regions, the algorithm performs the process described above at various scales in image segmentation and multiple combinations of diversification to measure the similarity of regions (for instance, CTSF, TSF, F, S). 
Therefore, objects of all sizes, shapes, and colors within the image are detected. 
A relevant drawback of this approach is the excessive number of candidate regions, thus generating a longer pre-processing time of the dataset images and the return of the search of a query.

\section{Proposed Method}

The proposed method aims at addressing several shortcomings of previous methods for PS and IR in historical documents. 
For instance, prior methods require a high computational effort and use a large amount of space to store feature maps, which leads to longer times to search for a query. Besides, the scalability of prior methods is difficult to achieve.
The proposed method detects object candidates in the document images with selection search algorithm and employs deep learning models for feature extraction. Such models consider either real-valued or binary code representations. 
Finally, candidate images are ranked by computing the feature similarity with a given input query.

\subsection{Object Detection with Selective Search (SS)}

An algorithm based on SS \cite{ss} was used for the object recognition task in the offline phase. However, this strategy presents difficulties when applied to images of historical documents. It generates several invalid regions, detecting antiquity stains, page edges, ink smudges, and other characteristics that do not represent objects of interest. Therefore, to minimize this problem, the SS algorithm was applied with only one scale variation using the Felzenszwalb and Huttenlocher algorithm \cite{felzen}, and only with the CTSF combination to measure the similarity of the regions. Despite considerably reducing the number of candidate regions compared to the method used in \cite{kelly}, several of the detected areas still represent invalid images. Thus, it was necessary to implement an algorithm that performs post-processing in the regions returned by SS.

The objective of filtering invalid regions is to exclude areas representing spots, background, page's edges, and very small or too large regions between the $x$ and $y$ axes. For filtering based on the regions' texture, a filter based on Gaussian derivation was used to highlight the edges in the image, which calculates the intensity of gradients in the image and reduces the potential edges to 1-pixel curves by removing non-maximum pixels from the gradient magnitude. Finally, the edge pixels are kept or removed using hysteresis thresholds on the magnitude of the gradient. As a result, we obtain a binary image with the object edges highlighted.

\begin{algorithm}[!b]
\footnotesize
    \SetKwData{BinaryImage}{binaryImage}
    \SetKwData{Sector}{sector}
    \SetKwData{Sectors}{sectors}
    \SetKwData{SectorsInvalid}{sectorsInvalid}
    \SetKwFunction{Mean}{Mean}
    \SetKwFunction{GetBinaryImage}{GetBinaryImage}
    \SetKwFunction{GetSectors}{GetSectors}
    \SetKwInOut{Input}{Input}\SetKwInOut{Output}{Output}
    \Input{A image $Img$ of dimension $h\times w$ of the candidate region and a threshold $\alpha$}
    \Output{Is the candidate region valid}
    \BlankLine
    \BinaryImage$\leftarrow$ \GetBinaryImage{$Img[h,w]$}\;
    \BlankLine
    \If{\Mean{\BinaryImage} $<$ $\alpha$}{
        \textbf{return} false;  /* \emph{candidate region is invalid}\ */ \\
    }
    \BlankLine
    /* \emph{generate eight sections from binary image}\ */ \\
    \Sectors$\leftarrow$ \GetSectors{\BinaryImage, 8}\;
    \SectorsInvalid$\leftarrow 0$\;
    \BlankLine
    \For{\Sector \textbf{in} \Sectors}{
        \If{\Mean{\Sector} $<$ $\alpha$}{
            \SectorsInvalid$\leftarrow$ \SectorsInvalid$+$ $1$\;  
        }
         \If{\SectorsInvalid $>$ $4$}{
            \textbf{return} false;  /* \emph{candidate region is invalid}\ */ \\
        }
    }
    \BlankLine
    \textbf{return} true;  /* \emph{candidate region is \textbf{valid}}\ */ \\
    \caption{Invalid candidate region filter}\label{invalidFilter}
\end{algorithm}

\begin{figure}[t]
    \centerline{\includegraphics[scale=0.45]{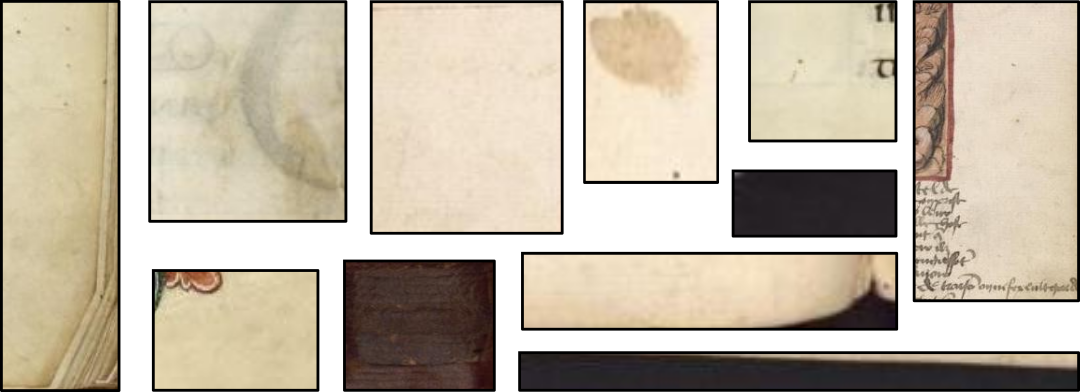}}
    \caption{Examples of regions filtered out by the algorithm.}
    \label{filter}
\end{figure}

Specific manipulations are performed on the binary images. First, the mean of the pixel values of the image is computed. Then, a minimum average of $\alpha$ is established as a parameter, so the image is immediately excluded if the number of points is less than $\alpha$. However, some invalid images may still be undetected only with the mean threshold. Therefore, a second step handles this problem by segmenting the image into eight sectors and averaging each sector individually. Then,  an image is also excluded if more than 50\% of the sectors had an average lower than $\alpha$. The pseudo-code of the invalid candidate region filter is shown in Algorithm~\ref{invalidFilter}. Such a filter reduces up to 1/5 the number of candidate regions returned by the SS without any training on the context images. Training was avoided so as not to affect the generalization of the proposed method if used on another image database. Fig.~\ref{filter} shows some examples of filtered invalid regions.

Finally, features are extracted from the valid candidate regions and stored to be used in the online phase.  As described in the next section, two CNN architectures were evaluated as base of the feature extractors.

\subsection{Feature Extraction}

A siamese neural network (SNN) is used to compute the similarity between the query and image candidates. SNN is a neural network architecture that combines two identical networks with the same configuration, parameters, and weights. Two images are used as input to the network to calculate their similarity. As a first approach, two CNN architectures were evaluated as alternatives to composing our SNN. The architectures used were the VGG19 \cite{vgg} and ResNet50V2 \cite{resnet50}. We chose these two architectures for their tried-and-true feature extraction abilities and their convenient availability in the deep learning framework in which the experiments were performed.

VGG19 \cite{vgg} is a CNN with 19 layers. This CNN is designed for applications in large-scale image classification and consists of five convolutional blocks interleaved with five max-pooling layers. Through some experiments in this work, it was possible to see great potential in applying this network as a feature extractor. Using the outputs of each block and concatenating them into a feature vector, we observed that some color and shape characteristics were well evidenced in the final result. We also evaluated the combination of pairs of blocks to find the best result.

ResNet50V2 is a residual deep network and emerged as a family of extremely deep architectures, having good precision and convergence \cite{resnet}. The base of ResNet50V2 is inspired by the philosophy of VGG \cite{vgg} networks. The convolutional layers use a 3$\times$3 filter and follow these rules: (i) for the same output feature map size; the layers have the same number of filters; and (ii) if the feature map size is reduced by half, the number of filters is doubled to preserve the per-layer complexity. Thus, ResNet50V2 has 50 convolution layers, max-pooling and average-pooling layers.

To apply the transfer learning concept, both VGG19 and ResNet50V2 were pre-trained in a supervised way on the ImageNet dataset \cite{imagenet}, with 1.28 million training samples and 50 thousand validation samples organized into 1000 classes, containing images from different contexts and objects. Several models of SNNs were built to establish a comparison between the two architectures and variations in the architectures themselves.

\begin{figure}[htbp]
    \centerline{\includegraphics[scale=0.35]{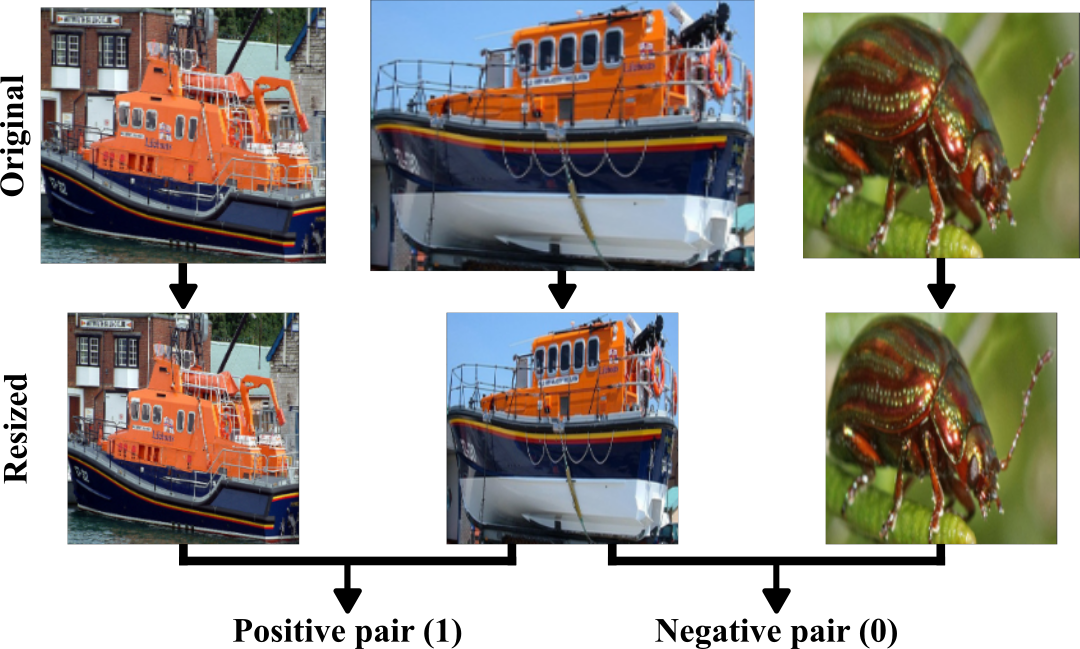}}
    \caption{Example of data preparation for training set creation using the Imagenet dataset.}
    \label{pairs}
\end{figure}

Pairs of images generated from the ImageNet dataset were used to train the SNNs. The images are resized to 224$\times$224 pixels using bilinear interpolation, as shown in Fig.~\ref{pairs}. Overall, 250,000 pairs were generated, where 150,000 are negative pairs, and 100,000 are positive pairs, following the ratio of 1.5$\times$ more negative pair images proposed by \cite{siamese}. 

\begin{figure}[htbp]
    \centerline{\includegraphics[scale=0.35]{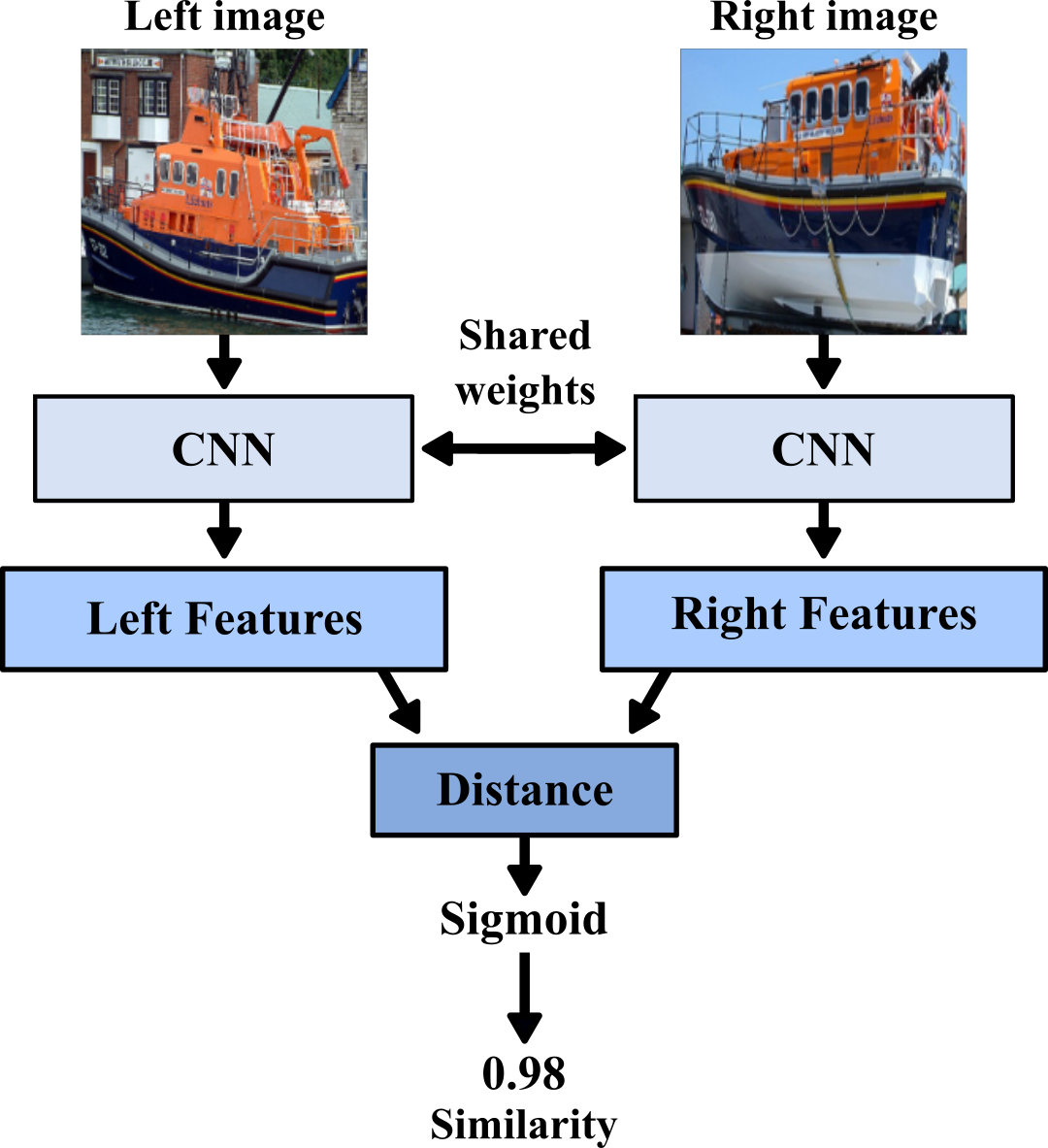}}
    \caption{Structure example of a Siamese Neural Network.}
    \label{siamesenet}
\end{figure}

The two images were used as input of the SNNs of the experiment. The Euclidean distance was calculated with the resulting feature maps to find the similarity. After the calculation, the distance measurement was normalized by a dense layer using sigmoid activation, as shown in Fig.~\ref{siamesenet}. Finally, the loss function used to train the network was the contrastive loss, proposed in \cite{contrastiveloss}.

For feature extraction with ResNet50V2, two variants of the network model were built. The first uses the last convolutional layer flatten, with 100,352 features, and the second uses the global average pooling layer, with 2048 features. The model using the convolutional layer was called \verb|ResNet Conv| and the model using the global average pooling layer was called \verb|ResNet GAPool|.

Four variants of the network were built for the feature extraction with the VGG19 architecture. As previously explained, the use of VGG19 block outputs is very sensitive to color and texture variations, something beneficial for the representation of features. To test which blocks have a better extraction, models were made using the concatenation of all blocks and the concatenation of pairs of these blocks. The first one using all blocks was called \verb|VGG19 Blocks|, containing 1472 features. The second one using blocks 4 and 5 was called \verb|VGG19 Block4-5|, with 1024 features. The third one using blocks 2 and 3 was called \verb|VGG19 Block2-3|, with 384 features. Moreover, the last one using blocks 2 and 5 was called \verb|VGG19 Block2-5|, containing 640 features.

With the creation of the models, they were used to construct the SNNs and trained with the pairs of positive and negative images. After training, the six models extracted features from all candidate images and their values stored in separate files. It is important to emphasize that the features were represented by floating-point values with 32 bits until that moment, representing a high storage cost.

To mitigate such a storage problem, deep hashing was also applied. The primary purpose of deep hashing is to transform the feature map with floating-point values into a more compact feature map using a binary code representation.

\subsection{Approach using Deep Hashing}

Hashing is one of the most used methods due to its efficiency in terms of computation and storage. It aims to convert the original high-dimensional features into low-dimensional hash codes so that the hash codes of similar objects are as close as possible and the hash codes of different objects are as diverse as possible. The purpose of the hashing algorithm is to map the features of the original space into a Hamming space, which leads us to compact hash codes composed of 0 and 1 values. As a result, hash encoding is very efficient in binary computation and storage, such as depicted in \cite{deephashing}.

The models previously built with deep learning were used in our hashing solution, thus composing deep hashing. A new layer was added at the end of each network. This layer consisted of transforming floating-point values into binary values, characterizing a discretization process. The construction of this layer was done using a discretization layer, which maps each element in a continuous interval. This algorithm was used considering a margin of value 1.

With the inclusion of this layer in each of the extractors of the SNNs, they were trained using the same process based on pairs of negative and positive images. However, the Hamming distance was the similarity calculation used for the hash codes. After training, the six models with hashing were used to extract the features of the candidate images, thus completing the offline phase.

\subsection{Similarity Calculations}

With the process of the offline phase completed, it is possible to use the system in the online phase. The online phase consists of extracting features from the query and comparing them with the entire list of candidates built in the offline phase. The features represented by floating-point values were compared using the Euclidean distance, while for the binary code features, the comparison was made with the Hamming distance, which consists of the sum of the absolute difference between each feature. However, when dealing with binary values, or values of only one bit, it is also possible to apply the XOR operation between the elements. In this operation, when comparing equal values, it returns 1, and with different values, it returns 0. The XOR operation can be done faster on the processor since it is a native bitwise operation. After comparing all the elements, a sum of all the results is made, as denoted in Equation \eqref{hamming}. 

\begin{equation}
    d=\sum\limits_{i=1}^n (q_i\ \textrm{XOR} \ p_i)\label{hamming}
\end{equation}

After the calculation, the results for each query were ordered for both the Euclidean and Hamming distances. Then, the lists of Top \(n\) candidates were generated to evaluate the models.

\section{Experiments}

The Experiments section is organized as follows. Subsection A presents the database used for testing and evaluating the proposed methods. Subsection B presents the results and improvements of selective search. Subsection C presents how the SNN models were trained and validated. Subsection D the results for the Image Retrieval task. Subsection E the results for the pattern spotting task. Finally, the processing time and storage cost results are presented in the last subsection.

\subsection{DocExplore}

The database used to carry out the experiments was DocExplore, proposed in \cite{docexplore} which consists of 1500 images of historical documents dating from the 10th to the 16th century. The original documents were extracted from the DocExplore project and are in possession of the Municipal Library of Rouen, France. All images were scanned at 600dpi resolution, resulting in dimensions ranging from 3000 pixels to 4000 pixels. Due to computational cost and storage, the images were compressed with a maximum of 1024 pixels in each dimension and 72dpi. The database provides 1447 unique queries, and their sizes range from 20$\times$11 pixels to 1307$\times$319 pixels.

\subsection{Selective Search}

The application of the SS algorithm in the DocExplore database was done with the diversification strategies combining the color element, texture, size, and fill, resulting in 976,486 candidate regions of objects. Through some experimentation, the value 0.06 was established for the $\alpha$ parameter of the invalid region filter function. After applying the invalid region filter, the number of candidate regions was dropped to 786,718, representing a decrease of 19.4\%. 

The number of candidate regions found by the algorithm used in Wiggers \textit{et al.}~\cite{kelly} was 36,159,870, approximately 46 times bigger than the result found in our work. This high number of candidate regions also represents a higher storage cost and processing time since the query must be compared with all the candidates.

\subsection{Siamese Neural Network}

For the training of the SNN models, the 250,000 pairs of images generated from the ImageNet database were used with a holdout of 70\% for training and 30\% for validation. Thus, the training set had 105,000 negative pairs and 70,000 positive pairs. The validation set had 45,000 negative pairs and 30,000 positive pairs. The training of all models were done with 25 epochs.

\subsection{Image Retrieval Task}

Tab.~\ref{resultsIR} shows the experimental results for the IR problem using the ResNet50V2 and VGG19 architectures applied to the set of candidate images returned by SS. The AlexNet network was also tested, the same architecture used by Wiggers \textit{et al.}~ \cite{kelly}. AlexNet was trained following the same steps described in Wiggers \textit{et al.}~ \cite{kelly} and applied to the same set of candidate images as ResNet50V2 and VGG19. The top 100, 300, 500, 700, and 1000 best results were evaluated for the IR task. In this task, the best result was in the top 1000 of VGG19 Block4-5 with an mAP of 53.21\%, surpassing by 10.4 percentage points the result of AlexNet and by 14.6 percentage points the result of Wiggers \textit{et al.}~ \cite{kelly}, which used selective search with 36,159,870 candidate images. These results depict that a filtered SS and returning a smaller number of candidates can positively influence the image retrieval result.

\begin{table}[htbp]
    \caption{Image Retrieval results}
    \begin{center}
        \begin{tabular}{lccccc}
            \toprule
            \multicolumn{6}{c}{\textbf{mAP for Image Retrieval}} \\ \hline
            \multicolumn{1}{c}{\multirow{2}{*}{\textbf{Method}}} & \multicolumn{5}{c}{\textbf{Top n}} \\ \cline{2-6} 
            & 100 & 300 & 500 & 700 & 1000 \\ 
            \hline
            ResNet Conv & 0.3956 & 0.4395 & 0.4567 & 0.4659 & 0.4723 \\ 
            ResNet GAPool & 0.4058 & 0.4617 & 0.4797 & 0.4880 & 0.4928 \\ 
            VGG19 Blocks & 0.3293 & 0.3934 & 0.4193 & 0.4279 & 0.4291 \\ 
            VGG19 Block4-5 & 0.4313 & 0.5058 & 0.5247 & 0.5307 & \textbf{0.5321} \\ 
            VGG19 Block2-3 & 0.3303 & 0.4060 & 0.4276 & 0.4346 & 0.4355 \\ 
            VGG19 Block2-5 & 0.4227 & 0.4939 & 0.5153 & 0.5217 & 0.5233 \\ 
            AlexNet & 0.3168 & 0.3996 & 0.4220 & 0.4271 & 0.4282 \\ 
            \bottomrule
        \end{tabular}
        \label{resultsIR}
    \end{center}
\end{table}

For testing the models using deep hashing, the two best networks of each of the ResNet50V2 and VGG19 architectures were chosen. The experimental protocol used was the same, only changing the similarity calculation for the Hamming distance. Tab.~\ref{resultsHashIR} shows the results for networks using deep hashing (H).

\begin{table}[htbp]
    \caption{Image Retrieval results with Hashing}
    \setlength{\tabcolsep}{4.5pt}
    \begin{center}
        \begin{tabular}{lccccc}
            \toprule
            \multicolumn{6}{c}{\textbf{mAP for Image Retrieval}} \\ \hline
            \multicolumn{1}{c}{\multirow{2}{*}{\textbf{Method}}} & \multicolumn{5}{c}{\textbf{Top n}} \\ \cline{2-6} 
            & 100 & 300 & 500 & 700 & 1000 \\ 
            \hline
            ResNet Conv H & 0.0754 & 0.0854 & 0.1012 & 0.1169 & 0.1257 \\ 
            ResNet GAPool H & 0.2945 & 0.3437 & 0.3632 & 0.3727 & 0.3784 \\ 
            VGG19 Block4-5 H & 0.3952 & 0.4564 & 0.4744 & 0.4822 & \textbf{0.4862} \\ 
            VGG19 Block2-5 H & 0.3302 & 0.3777 & 0.3939 & 0.4026 & 0.4090 \\ 
            \bottomrule
        \end{tabular}
        \label{resultsHashIR}
    \end{center}
\end{table}

Now comparing the experimental results against state-of-the-art approaches, we observe that the method proposed by En \textit{et al.}~ \cite{sovann} outperforms the VGG19 Block4-5 and VGG19 Block4-5 Hashing methods by 4.8 and 9.4 percentage points, respectively. However, an advantage of the methods proposed in this work is that at no time do they use information from the DocExplore database to refine their results. In Tab.~\ref{stateArtIR} we can visualize the main results of state-of-the-art in comparison with the best results of this work.

\begin{table}[htbp]
    \caption{Comparison of the methods with the state-of-the-art IR}
    \begin{center}
        \begin{tabular}{lc}
            \toprule
            \multicolumn{1}{c}{Methods} & IR Top 1000 \\ 
            \hline
            En \textit{et al.}~ \cite{sovann} & \textbf{0.580} \\
            Ubeda \textit{et al.}~ \cite{ignacio} ES & 0.286 \\
            Ubeda \textit{et al.}~ \cite{ignacio} PP & 0.386 \\
            Wiggers \textit{et al.}~ \cite{kelly} PP & 0.386 \\
            VGG19 Block4-5 & 0.532 \\
            VGG19 Block4-5 H & 0.486 \\
            \bottomrule
        \end{tabular}
        \label{stateArtIR}
    \end{center}
\end{table}

\subsection{Pattern Spotting Task}

For the PS task, the mAP was evaluated taking into account an IoU $\geq$ 0.5 and with the 100, 300, 500, 700, and 1000 best similarity results.

Tab.~\ref{resultsPS} shows the experimental results for the PS problem. As in the image retrieval, the two best networks of each of the ResNet50V2 and VGG19 architectures were chosen for testing the models using deep hashing.

\begin{table}[htbp]
    \caption{Pattern Spotting results}
    \begin{center}
        \begin{tabular}{lccccc}
            \toprule
            \multicolumn{6}{c}{\textbf{mAP for Pattern Spotting}} \\ \hline
            \multicolumn{1}{c}{\multirow{2}{*}{\textbf{Method}}} & \multicolumn{5}{c}{\textbf{Top n}} \\ \cline{2-6} 
            & 100 & 300 & 500 & 700 & 1000 \\ 
            \hline
            ResNet Conv & 0.1447 & 0.1705 & 0.1738 & 0.1751 & \textbf{0.1761} \\ 
            ResNet GAPool & 0.1225 & 0.1478 & 0.1524 & 0.1542 & 0.1557 \\ 
            VGG19 Blocks & 0.0724 & 0.0848 & 0.0876 & 0.0888 & 0.0898 \\ 
            VGG19 Block4-5 & 0.0997 & 0.1196 & 0.1237 & 0.1254 & 0.1268 \\ 
            VGG19 Block2-3 & 0.0643 & 0.0761 & 0.0795 & 0.0811 & 0.0825 \\ 
            VGG19 Block2-5 & 0.1118 & 0.1339 & 0.1386 & 0.1407 & 0.1425 \\ 
            AlexNet & 0.0610 & 0.0674 & 0.0689 & 0.0697 & 0.0703 \\  
            \bottomrule
        \end{tabular}
        \label{resultsPS}
    \end{center}
\end{table}

 Tab.~\ref{resultsHashPS} shows the results for networks using deep hashing (H). Contrary to some expectations, the network using the VGG19 Block4-5 performed better in binary code than in floating-point values.

\begin{table}[htbp]
    \caption{Pattern Spotting results with Hashing}
    \setlength{\tabcolsep}{4.5pt}
    \begin{center}
        \begin{tabular}{lccccc}
            \toprule
            \multicolumn{6}{c}{\textbf{mAP for Pattern Spotting}} \\ \hline
            \multicolumn{1}{c}{\multirow{2}{*}{\textbf{Method}}} & \multicolumn{5}{c}{\textbf{Top n}} \\ \cline{2-6} 
            & 100 & 300 & 500 & 700 & 1000 \\ 
            \hline
            ResNet Conv H & 0.0303 & 0.0311 & 0.0313 & 0.0314 & 0.0315 \\ 
            ResNet GAPool H & 0.0911 & 0.1040 & 0.1061 & 0.1070 & 0.1077 \\ 
            VGG19 Block4-5 H & 0.1094 & 0.1341 & 0.1388 & 0.1409 & \textbf{0.1426} \\ 
            VGG19 Block2-5 H & 0.0935 & 0.1129 & 0.1163 & 0.1176 & 0.1186 \\ 
            \bottomrule
        \end{tabular}
        \label{resultsHashPS}
    \end{center}
\end{table}

With En \textit{et al.}~ \cite{sovann} and Wiggers \textit{et al.}~ \cite{kelly}, it was observed that multiple candidate images often contained only part of the query or overlapped with other candidate images, thus reducing system performance. To alleviate the aforementioned problem, a post-processing step was proposed to use a union of these retained candidate images to discover rectangular regions as a way to improve the performance of the Pattern Spotting task. Thus, the first 3000 candidates were selected, and the union step was applied, assuming an IoU of 0.85. If two images had an IoU measurement greater than 0.85, the image with the smallest distance was kept, and the other was excluded. After the union, the first 100, 300, 500, 700, and 1000 were considered to feed the evaluation system. For all results, an improvement in mAP was observed after post-processing (PP).

\begin{table}[htbp]
    \caption{Pattern Spotting results with PP}
    \setlength{\tabcolsep}{4.5pt}
    \begin{center}
        \begin{tabular}{lccccc}
            \toprule
            \multicolumn{6}{c}{\textbf{mAP for Pattern Spotting}} \\ \hline
            \multicolumn{1}{c}{\multirow{2}{*}{\textbf{Method}}} & \multicolumn{5}{c}{\textbf{Top n}} \\ \cline{2-6} 
            & 100 & 300 & 500 & 700 & 1000 \\ 
            \hline
            ResNet Conv PP & 0.1716 & 0.1946 & 0.1974 & 0.1986 & \textbf{0.1996} \\ 
            ResNet Conv H PP & 0.0332 & 0.0340 & 0.0342 & 0.0344 & 0.0345 \\ 
            ResNet GAPool PP & 0.1432 & 0.1674 & 0.1712 & 0.1729 & 0.1743 \\ 
            ResNet GAPool H PP & 0.1059 & 0.1173 & 0.1192 & 0.1200 & 0.1207 \\ 
            VGG19 Blocks PP & 0.0861 & 0.0975 & 0.1000 & 0.1011 & 0.1021 \\ 
            VGG19 Block4-5 PP & 0.1181 & 0.1364 & 0.1401 & 0.1417 & 0.1429 \\ 
            VGG19 Block4-5 H PP & 0.1302 & 0.1531 & 0.1572 & 0.1593 & 0.1610 \\ 
            VGG19 Block2-3 PP & 0.0755 & 0.0867 & 0.0899 & 0.0915 & 0.0927 \\ 
            VGG19 Block2-5 PP & 0.1331 & 0.1535 & 0.1577 & 0.1598 & 0.1615 \\ 
            VGG19 Block2-5 H PP & 0.1119 & 0.1294 & 0.1322 & 0.1335 & 0.1345 \\ 
            AlexNet PP & 0.0691 & 0.0753 & 0.0768 & 0.0775 & 0.0781 \\ 
            \bottomrule
        \end{tabular}
        \label{resultsPSPP}
    \end{center}
\end{table}

In comparison with the results against state-of-the-art methods, we have that the ResNet Conv PP method surpasses the one proposed by Wiggers \textit{et al.}~ \cite{kelly} PP at 2.56 percentage points. In Tab.~\ref{stateArtPS} we can visualize the main results of state of art compared to the models presented in this paper.

\begin{table}[htbp]
    \caption{Comparison of the methods with the state-of-the-art PS}
    \begin{center}
        \begin{tabular}{lc}
            \toprule
            \multicolumn{1}{c}{Method} & PS Top 1000\\ 
            \hline
            En \textit{et al.}~ \cite{sovann} & 0.1570 \\
            Ubeda \textit{et al.}~ \cite{ignacio} ES & 0.1390 \\
            Ubeda \textit{et al.}~ \cite{ignacio} PP & 0.1730 \\
            Wiggers \textit{et al.}~ \cite{kelly} PP & 0.1740 \\
            ResNet Conv PP & \textbf{0.1996} \\
            VGG19 Block4-5 H PP & \textbf{0.1610} \\ 
            \bottomrule
        \end{tabular}
        \label{stateArtPS}
    \end{center}
\end{table}



\begin{figure}[ht]
    \centerline{\includegraphics[scale=0.3]{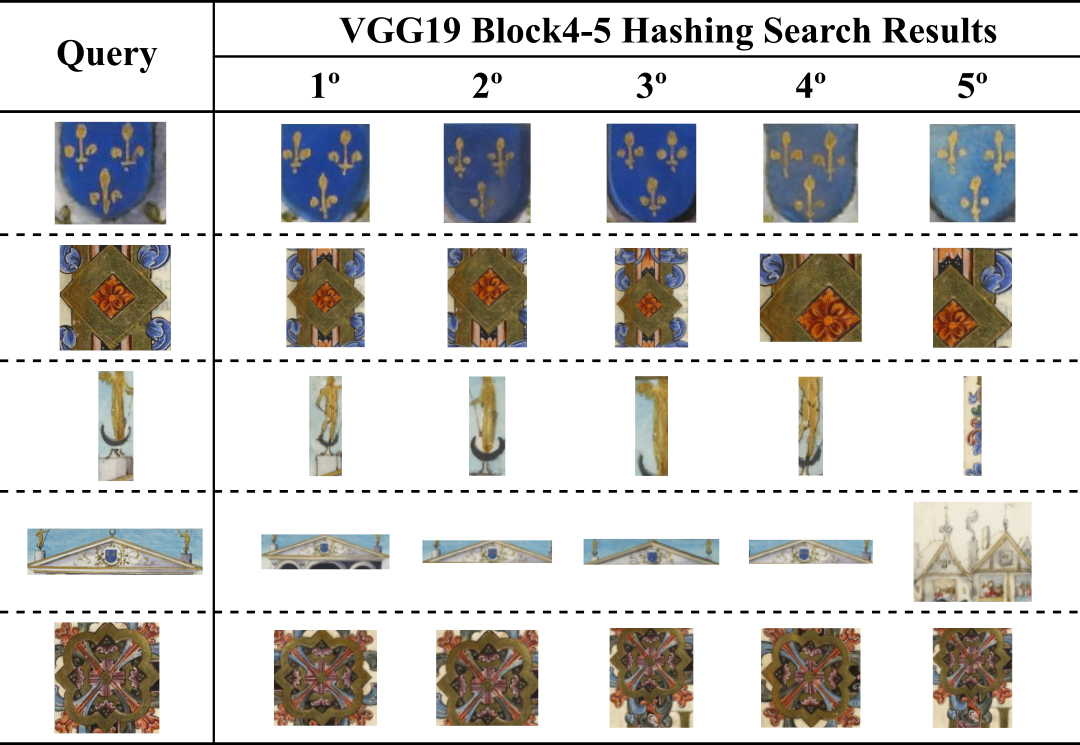}}
    \caption{Qualitative results of the search of some queries in the DocExplore database. The figure shows the image used in the query and its first five results returned by the VGG19 Block4-5 Hashing PP method.}
    \label{resultVGG}
\end{figure}

Fig.~\ref{resultVGG} shows the qualitative results of the images returned for the search of five queries using the VGG19 Block4-5 Hashing feature map, with 1024 dimensions. These results are promising, considering that most of the Top 5 images are similar to those used in the searches and the feature extractor used was based on deep hashing.

\subsection{Search Time and Storage Cost}

Tab.~\ref{resultsTimeStore} shows the average time required for query search and the amount of storage needed to save the feature maps of all candidates in the offline phase. Time is averaged over 50 different queries. The storage space considers only the feature map's size, without any additional structure to facilitate indexing and storage.

\begin{table}[htbp]
    \caption{Results for processing time and storage}
    \setlength{\tabcolsep}{4pt}
    \begin{tabular}{lrrcrc}
        \toprule
        \multicolumn{1}{c}{\multirow{2}{*}{\textbf{Method}}} & \multicolumn{1}{c}{\multirow{2}{*}{\textbf{Features}}} & \multicolumn{2}{c}{\textbf{Time (s)}} & \multicolumn{2}{c}{\textbf{Storage (GB)}} \\ \cline{3-6} 
        \multicolumn{1}{c}{} & \multicolumn{1}{c}{} & \multicolumn{1}{c}{\textbf{FP}} & \textbf{Binary} & \multicolumn{1}{c}{\textbf{FP}} & \textbf{Binary} \\ \hline
        ResNet Conv & 100 352 & \multicolumn{1}{r}{44.59} & \multicolumn{1}{r}{20.75} & \multicolumn{1}{r}{294.11} & \multicolumn{1}{r}{9.19} \\ \hline
        ResNet GAPool & 2 048 & \multicolumn{1}{r}{4.54} & \multicolumn{1}{r}{3.07} & \multicolumn{1}{r}{6.00} & \multicolumn{1}{r}{0.19} \\ \hline
        VGG19 Blocks & 1 472 & \multicolumn{1}{r}{4.19} & --- & \multicolumn{1}{r}{4.31} & --- \\ \hline
        VGG19 Block4-5 & 1 024 & \multicolumn{1}{r}{4.10} & \multicolumn{1}{r}{2.94} & \multicolumn{1}{r}{3.00} & \multicolumn{1}{r}{0.09} \\ \hline
        VGG19 Block2-3 & 384 & \multicolumn{1}{r}{3.91} & --- & \multicolumn{1}{r}{1.13} & --- \\ \hline
        VGG19 Block2-5 & 640 & \multicolumn{1}{r}{3.94} & \multicolumn{1}{r}{2.85} & \multicolumn{1}{r}{1.88} & \multicolumn{1}{r}{0.06} \\ \hline
        AlexNet & 4 096 & \multicolumn{1}{r}{12.81} & --- & \multicolumn{1}{r}{12.00} & --- \\ \hline
        Wiggers \textit{et al.} \cite{kelly} & 4 096 & \multicolumn{1}{r}{588.65} & --- & \multicolumn{1}{r}{551.76} & --- \\
        \bottomrule
        \multicolumn{6}{l}{FP: Floating-point.}
    \end{tabular}
    \label{resultsTimeStore}
\end{table}

\subsection{Discussion}

The results show that the models using the VGG19 architecture were better in the Image Retrieval task, where it is necessary to find only the first best occurrence of the query on the page. However, for the Pattern Spotting task, where it is required to compose a list of similar images and locate them on the page, the models using the ResNet50V2 architecture achieved a better result.

The significant improvement presented in this work was reducing the computational effort applied to solve the problem. 
While in \cite{kelly}, the candidate regions were more than 36 million, the proposed method achieved approximately 780 thousand candidate regions. Consequently, it was possible to optimize the results and the processing time both in the IR task and in the PS task, as observed in Tables~\ref{stateArtPS},~\ref{stateArtIR} and~\ref{resultsTimeStore}. 
While the method proposed by Wiggers \textit{et al.} \cite{kelly} needed more than 580 seconds to search for a query, the proposed method performs the same search in a maximum of 45 seconds while achieving superior results.

Another advantage was the application of deep hashing. Even transforming the characteristics to binary values, the result was not so harmed when we take into account that in the method proposed by Wiggers \textit{et al.} \cite{kelly}, the search for a query could take 200 times longer than Hashing methods. In addition to the search time, the cost in terms of storage was also significantly reduced. In the method proposed by Wiggers \textit{et al.} \cite{kelly}, more than 550 GB would be needed, while in the approach using VGG19 with Hashing, this number is no more than 0.09 GB.

It was possible to notice that in some cases, the transformation of the characteristics to the binary domain can increase the result, as observed in the VGG19 Block4-5 model where we had an improvement of almost 2 percentage points. Regarding query search time, converting to binary code reduced the time by approximately 30\% for methods using VGG19. In terms of storage, we migrated from a structure of floating-point values with 32 bits and started to use only 1 bit, reducing the overall storage cost by 32 times.

\section{Conclusion}

This paper presented two approaches for the IR and PS tasks for images of a historical document collection. The first approach improved an existing method reducing the number of candidate images returned by the selective search and improving the mAP, processing time, and storage cost. However, we could observe that although the IR task is part of the PS task, they will not necessarily present the same performance gains over different CNN architectures.

The second approach reduced both storage cost and processing time by deep hashing techniques. Nevertheless, the significant contribution of this investigation was to observe that in some cases, the conversion to the binary domain can also increase the result achieved with real-valued features. In cases where this does not happen, the performance loss is mitigated by reducing computational resources usage.

Future work will focus on fine-tuning the CNNs used as feature extractors images of historical documents since such CNNs were only trained on the ImageNet dataset. We believe that the mAP could be improved if such CNNs were fine-tuned on images of a similar context. Besides, replacing the selective search with a dynamic search strategy may also improve retrieval performance. 

\balance
\bibliographystyle{IEEEtranS}
\bibliography{IEEEreferences}
\balance

\end{document}